\documentclass[runningheads]{llncs}
\usepackage[T1]{fontenc}
\usepackage{lmodern}

\usepackage{graphicx}

\usepackage{enumitem}

\usepackage{chessfss}

\usepackage{comment}

\usepackage{xcolor}

\usepackage{geometry}

\usepackage{hyperref}

\usepackage{cite}

\usepackage{amsmath}

\usepackage{cleveref}

\begin{document}

\newcommand{\AL}[1]{\underline{#1}}

\title
{PAWN}

\subtitle
{
\AL{P}iece Value \AL{A}nalysis \AL{w}ith \AL{N}eural Networks
}

\author{Ethan Tang\inst{1}\orcidID{0009-0009-7901-7094} \and \\
Hasan Davulcu\inst{1}\orcidID{0000-0001-5602-8270} \and 
\\
Jia Zou\inst{1}\orcidID{0000-0001-9849-0711} \and
\\
Zhongju Zhang\inst{1}\orcidID{0000-0001-9200-2369}}
\authorrunning{E. Tang et al.}

\institute{Arizona State University, Tempe AZ 85281, USA\\
\email{\{ejtang,hdavulcu,jia.zou,zhongju.zhang\}@asu.edu}}

\maketitle          

\begin{abstract}
Predicting the relative value of any given chess piece in a position remains an open challenge, as a piece's contribution depends on its spatial relationships with every other piece on the board. We demonstrate that incorporating the state of the full chess board via latent position representations derived using a CNN-based autoencoder significantly improves accuracy for MLP-based piece value prediction architectures. Using a dataset of over 12 million piece-value pairs gathered from Grandmaster-level games, with ground-truth labels generated by Stockfish 17, our enhanced piece value predictor significantly outperforms context-independent MLP-based systems, reducing validation mean absolute error by 16\% and predicting relative piece value within approximately 0.65 pawns. More generally, our findings suggest that encoding the full problem state as context provides useful inductive bias for predicting the contribution of any individual component.


\end{abstract}

\section{Introduction}
\label{sec:background}

\subsection{Background} 
\label{sec:background_intro}

Chess has remained a topical area of research for AI systems, with chess transitioning into a ``digital period'' starting with Deep Blue's defeat of the then-reigning World Chess Champion Grandmaster (GM) Garry Kasparov in their 1997 rematch \cite{ref_deepblue}. Traditional chess engine architectures followed Deep Blue's human-designed position evaluation heuristics for the next decade, further increasing the gap in playing strength between humans and computers. New improvements revolutionized chess engine design in the late-2010s, with DeepMind's AlphaZero dethroning Stockfish \cite{ref_alphazero}, then widely regarded as the strongest chess engine, by combining reinforcement learning with deep neural networks in place of Stockfish's handcrafted evaluation function and alpha-beta search. Efficiently updatable neural networks (NNUE), introduced for computer shogi \cite{ref_nnue_nasu,ref_nnue_translation} in 2018 and subsequently adapted to chess, brought learned evaluation into alpha-beta search, allowing Stockfish 10 to reclaim its position as the strongest chess engine in 2020. The cumulative effect of these innovations has been dramatic. The original version of Stockfish, with an estimated Elo of 2747 \cite{ref_ccrl}, would be expected to win zero games in a 100-game match against its newest descendant Stockfish 18, rated at 3692 \cite{ref_sf18, ref_elo}.

\subsection{Traditional Chess Piece Valuation Systems}
\label{sec:background_trad_cp_val}

While these advances have dramatically improved chess engines' ability to select strong moves, the evaluations they produce remain opaque. An engine may correctly judge that exchanging a bad knight for a good bishop improves the position, but extracting why from its evaluation is nontrivial. A closely related and more interpretable line of research concerns the value of individual chess pieces, which underpins core strategic decisions such as when to initiate exchanges and how to assess material imbalances. Throughout a rich history spanning back to the 18th century, numerous general piece valuation systems have been proposed. General piece valuation systems typically express the values of each piece type as ratios, with single pawns as a base unit of measurement. Beginners to chess are likely familiar with the system of ({\raisebox{-0.3ex}{\resizebox{10pt}{!}{\WhitePawnOnWhite}}}=1,
{\raisebox{-0.3ex}{\resizebox{10pt}{!}{\WhiteKnightOnWhite}}}=3,
{\raisebox{-0.3ex}{\resizebox{10pt}{!}{\WhiteBishopOnWhite}}}=3,
{\raisebox{-0.3ex}{\resizebox{10pt}{!}{\WhiteRookOnWhite}}}=5,
{\raisebox{-0.3ex}{\resizebox{10pt}{!}{\WhiteQueenOnWhite}}}=9)
\cite{ref_piece_values_wiki}, defined by the Modenese School in the 18th century. Later systems in the 20th century by Lasker \cite[p.~73]{ref_lasker}, Turing \cite{ref_turing}, and Fischer \cite[p.~14]{ref_fischer}
 largely preserved this structure but
diverged on key points: Turing and
Fischer gave the value of bishops ({\raisebox{-0.3ex}{\resizebox{10pt}{!}{\WhiteBishopOnWhite}}}) a
slight edge over knights ({\raisebox{-0.3ex}{\resizebox{10pt}{!}{\WhiteKnightOnWhite}}}) at 3.5/3.25 vs. 3 pawns respectively, while opinions on the value of queens
({\raisebox{-0.3ex}{\resizebox{10pt}{!}{\WhiteQueenOnWhite}}}) ranged from 9 to
10. Both Lasker and Fischer also addressed the value of kings ({\raisebox{-0.3ex}{\resizebox{10pt}{!}{\WhiteKingOnWhite}}}), a question the
Modenese system left open --- Lasker assigned it a value of 4, while
Fischer treated it as invaluable ($\infty$).

\subsection{Contemporary Chess Piece Valuation Systems}
\label{sec:background_new_cp_val}

More contemporary approaches to developing general piece valuation systems have utilized additional features including material difference or game stage. DeepMind (2020) utilized the relative effect of piece counts in a given position on AlphaZero's predicted game outcomes to derive piece values of ({\raisebox{-0.3ex}{\resizebox{10pt}{!}{\WhitePawnOnWhite}}}=1, {\raisebox{-0.3ex}{\resizebox{10pt}{!}{\WhiteKnightOnWhite}}}=3.05, {\raisebox{-0.3ex}{\resizebox{10pt}{!}{\WhiteBishopOnWhite}}}=3.33, {\raisebox{-0.3ex}{\resizebox{10pt}{!}{\WhiteRookOnWhite}}}=5.63, {\raisebox{-0.3ex}{\resizebox{10pt}{!}{\WhiteQueenOnWhite}}}=9.5) \cite{ref_alphazero_balance}. Similarly, GM Larry Kaufman (2022) used additional contextual features including game stage (middlegame vs. threshold vs. endgame), pawn file, or bishop pair (whether only one side has two bishops) in his piece valuation system\footnote{Kaufman's system assigns different general piece values depending on whether queens are present or not on the board. The values shown here are his baseline middlegame values with queens present \cite{ref_piece_values_wiki}.} of ({\raisebox{-0.3ex}{\resizebox{10pt}{!}{\WhitePawnOnWhite}}}=1, {\raisebox{-0.3ex}{\resizebox{10pt}{!}{\WhiteKnightOnWhite}}}=3.25, {\raisebox{-0.3ex}{\resizebox{10pt}{!}{\WhiteBishopOnWhite}}}=3.5, {\raisebox{-0.3ex}{\resizebox{10pt}{!}{\WhiteRookOnWhite}}}=5, {\raisebox{-0.3ex}{\resizebox{10pt}{!}{\WhiteQueenOnWhite}}}=9.75) \cite{ref_kaufman_chesscom,ref_kaufman_imbalances}. 

The most recent approaches to piece valuation incorporate data-driven machine-learning methods to derive independent interpretations of piece value. Gupta et al. (2023) trained a multi-layer perceptron (MLP) to predict expected game outcomes for (Color, Piece, Square) triplets using a dataset of Grandmaster games, creating heatmaps of ideal squares for pawns, knights, and bishops \cite{ref_gupta}. Pav (2025) applied methods similar to DeepMind's AlphaZero-derived general piece valuation system, using logistic regression on a large sample of online games to validate other traditional piece valuation systems \cite{ref_pav}. Spinnato (2025) utilized SHapley Additive exPlanations (SHAP) to explain engine evaluations through piece ablation, defining a piece's contribution to a position as the change in engine evaluation between a position with or without the piece \cite{ref_shap}. 

\subsection{Our Contribution}
\label{sec:background_our_contribution}

Our paper connects and builds on several recent works utilizing machine learning to examine piece valuation systems. We adopt Spinnato's ablation-based definition of piece value as the change in the evaluation of a position based on a piece's removal. Building on Gupta's MLP architecture, we use a convolutional neural network (CNN) based autoencoder to derive latent representations of the board state and append them as additional context for MLP-based piece value prediction. Additionally, we extract a larger and more robust dataset of Grandmaster-level games and ground-truth piece value labels using Stockfish 17, along with extending piece value predictions to all non-king piece types. Finally, we demonstrate that including the immediate board state using a CNN-autoencoder-derived latent representation significantly increases the accuracy of piece value predictions for MLP-based systems. We also highlight the limitations of static piece valuation systems and explore further avenues of improvement regarding predictive systems for relative chess piece value.

\section{Problem Statement}
\label{sec:problem}

\subsection{What is Chess Piece Value?}
\label{sec:problem_pval_def_comparison}

We first examine the difference between the general and relative value of a chess piece. General value refers to the static piece valuation systems covered in Sections \ref{sec:background_trad_cp_val} and \ref{sec:background_new_cp_val}, where each piece type is assigned a fixed value using pawns as a base unit. The general value of a piece represents its average, context-free worth across all valid chess positions. In contrast, relative value captures a piece's contribution within a specific position, dependent on its relationships to both enemy and ally pieces. GM Rowson refers to relative piece value as ``Quality'' in his book `Chess for Zebras', noting that it encompasses ``everything from weak squares, vulnerable pawns, strong doubled pawns...[to] elusive ideas like `coordination' and `harmony'\,'' \cite[p.~116]{ref_rowson}. In this paper, we focus on predicting the relative value of a piece in any given position.

\begin{figure}[htbp]
\centering
\includegraphics[width=0.50\textwidth]{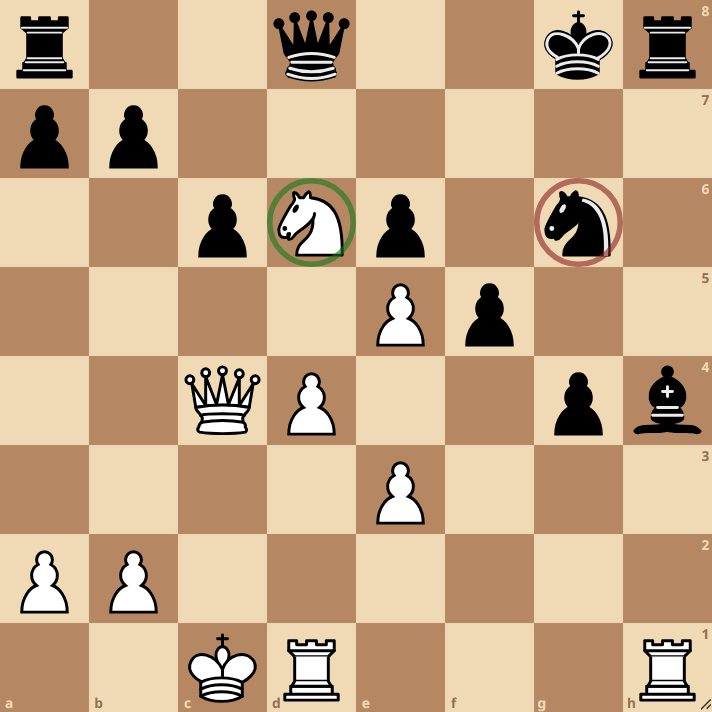}
\caption{In this position \cite{ref_artemiev_rozum} with White to move, our piece value predictor assigned the {\raisebox{-0.3ex}{\resizebox{9pt}{!}{\WhiteKnightOnWhite}}}d6 a piece value of 703 cp, which is significantly larger than the {\raisebox{-0.3ex}{\resizebox{9pt}{!}{\BlackKnightOnWhite}}}g6 assigned a piece value of -355 cp. 
}
\label{fig1}
\end{figure}

Figure \ref{fig1} illustrates how a piece's relative value depends on its relation to both enemy and ally pieces. The {\raisebox{-0.3ex}{\resizebox{10pt}{!}{\WhiteKnightOnWhite}}}d6 is significantly more valuable than its counterpart on g6, acting as an important contributor to White's attack on Black's {\raisebox{-0.3ex}{\resizebox{10pt}{!}{\BlackKingOnWhite}}}g8 via the f7 square. Meanwhile, the {\raisebox{-0.3ex}{\resizebox{10pt}{!}{\BlackKnightOnWhite}}}g6 is far from its ideal square of d5 and is unable to make it there before White cracks open Black's structure with a devastating attack following 22. {\raisebox{-0.3ex}{\resizebox{10pt}{!}{\WhiteQueenOnWhite}}}xe6+ {\raisebox{-0.3ex}{\resizebox{10pt}{!}{\BlackKingOnWhite}}}h7 24. {\raisebox{-0.3ex}{\resizebox{10pt}{!}{\WhiteKnightOnWhite}}}xf5. Black resigned immediately after 24. {\raisebox{-0.3ex}{\resizebox{10pt}{!}{\WhiteKnightOnWhite}}}xf5, with Stockfish 18 giving mate-in-10 for White.

\subsection{Formal Chess Piece Value Definition}
\label{sec:problem_formal_pval_def}

\noindent We define the relative value of a piece $v_{\text{piece}}$, also known as ``piece quality,'' as:

\begin{equation}
\label{eq1}
v_{\text{x}} = E(P) - E(P \setminus x)
\end{equation}

\noindent where:

\begin{itemize}[leftmargin=4em]
\item[$v_{\text{x}}$] is the relative value of piece $x$ in position $P$
\item[$E(P)$] is the engine evaluation of position $P$
\item[$E(P \setminus x)$] is the engine evaluation of position $P$ with piece $x$ removed
\end{itemize}

\begin{figure}[ht]
\centering
\includegraphics[width=0.50\textwidth]{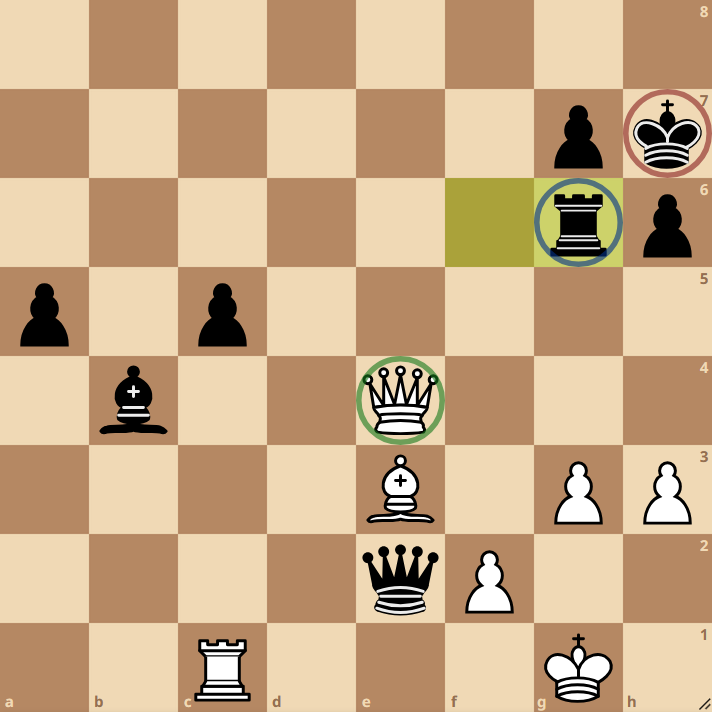}
\caption{In this position \cite{ref_ding_nepo} with White to move, the piece value of the {\raisebox{-0.3ex}{\resizebox{9pt}{!}{\BlackRookOnWhite}}}g6 cannot be calculated using our definition since the {\raisebox{-0.3ex}{\resizebox{9pt}{!}{\WhiteQueenOnWhite}}}e4 would attack the {\raisebox{-0.3ex}{\resizebox{9pt}{!}{\BlackKingOnWhite}}}h7.} \label{fig2}
\label{fig:invalid_pval_ex}
\end{figure}

Chess engines evaluate positions in units of centipawns (cp) \cite{ref_centipawns}, which approximate the advantage a given side has in hundredths of pawns (e.g. an evaluation of +200 cp means White has an advantage equivalent to roughly two extra pawns). Our definition of piece value captures a piece's relative contribution to the overall evaluation of a position by measuring the quantitative impact of its removal. Kings are excluded, as removing a king produces an illegal position under our definition. Additionally, specific pieces of interest are skipped in cases where removing the piece yields an illegal position (such as in Fig. \ref{fig:invalid_pval_ex}).

\section{System Architecture}
\label{sec:sysarch}

\subsection{Overview}
\label{sec:sysarch_overview}

Computing $v_{\text{x}}$ for a single piece requires two separate engine evaluations: one for the original position and one for the position with the piece removed. Across all non-king pieces in a given position, inference quickly becomes expensive and slow. Our system instead learns to predict $v_{\text{x}}$ directly from the board state, avoiding repeated engine calls. Building on Gupta's MLP predictor \cite{ref_gupta}, we incorporate the full board state as additional context alongside the original piece location input of (Piece, Color, Square). To evaluate the impact of this enhancement, we trained two categories of piece value predictors: a set of baseline multi-layer perceptron models (referred to as MLP) and an enhanced variant that augments the MLP input with a latent representation of the board state generated by a convolutional neural network autoencoder (referred to as MLP+CNN).

Both MLP and MLP+CNN models share a common input representation for individual pieces, encoding piece type as a one-hot vector and piece location as a normalized coordinate vector. The MLP+CNN models additionally receive an intermediate representation of the entire board state, extracted by a CNN autoencoder from a $12 \times 8 \times 8$ binary embedding of the position, where each of the 12 piece-type channels encodes the presence (1) or absence (0) of that piece type on each of the 64 squares. Sections \ref{sec:sysarch_mlp_pval_pred} and \ref{sec:sysarch_mlp_cnn_pval_pred} detail the specific input dimensions and architectural choices for each model category.

\subsection{MLP Piece Value Predictors}
\label{sec:sysarch_mlp_pval_pred}

Three distinct models of MLP-based piece value predictors were trained as baselines:
\begin{itemize}
\item MLP \#1: A recreation of Gupta's 2-layer MLP with hidden layers [64, 32].
\item MLP \#2: A 3-layer MLP with hidden layers [128, 64, 32].
\item MLP \#3: A 3-layer MLP with hidden layers [128, 64, 32] using additional input features of rank$^2$ and file$^2$.
\end{itemize}

MLP \#1 and \#2 take a 12-dimensional input: a 10-dim one-hot encoding over White/Black non-king piece types concatenated with a 2-dim location vector for rank and file. MLP \#3 takes an augmented 14-dimensional input, appending rank$^2$ and file$^2$ terms to capture non-linear positional patterns (e.g. passed pawns, knights on the rim). All coordinate features are normalized to $[0, 1]$ by dividing by 7 prior to training. Each MLP hidden layer applies a linear transformation followed by batch normalization, ReLU activation, and dropout ($p=0.2$). A final linear layer maps the last hidden layer's output to a single scalar, representing the predicted value of the input piece in the given position.

\subsection{MLP+CNN Piece Value Predictors}
\label{sec:sysarch_mlp_cnn_pval_pred}

\paragraph {Motivating example.} Our MLP+CNN models augment the baseline MLP input with a latent representation of the full board state produced by a CNN autoencoder. By appending a $d$-dimensional position representation to the 14-dimensional piece feature vector used by MLP \#3, our enhanced models receive both local piece information and global positional context.

\begin{figure}[ht]
\centering
\includegraphics[width=0.45\textwidth]{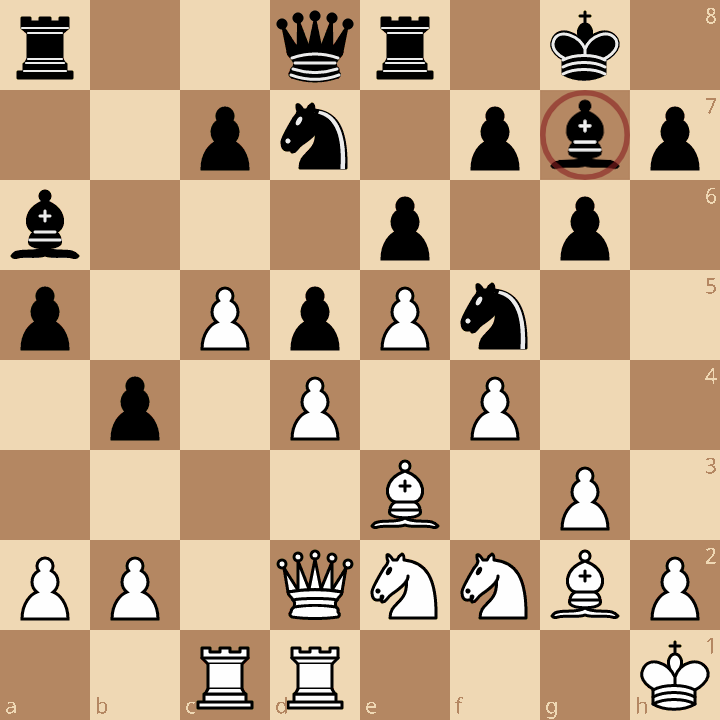}\hspace{0.75em}
\includegraphics[width=0.45\textwidth]{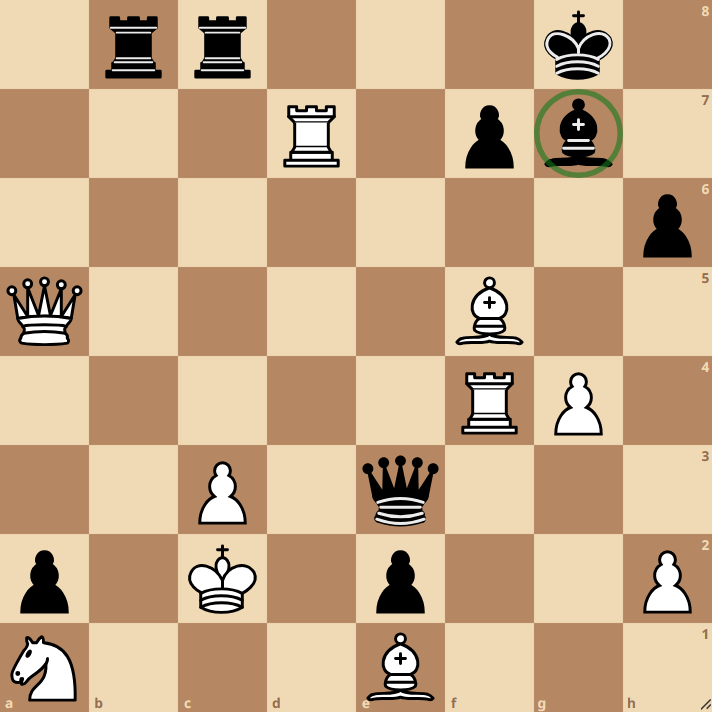}
\caption{Our piece value predictor assigns the bad {\raisebox{-0.3ex}{\resizebox{9pt}{!}{\BlackBishopOnWhite}}}g7 (left) biting on White's granite of {\raisebox{-0.3ex}{\resizebox{9pt}{!}{\WhitePawnOnWhite}}}d4-e5-f4 a modest piece value of -453 cp. Meanwhile, the active {\raisebox{-0.3ex}{\resizebox{9pt}{!}{\BlackBishopOnWhite}}}g7 (right), which acts as a key contributor to Black's attack on the {\raisebox{-0.3ex}{\resizebox{9pt}{!}{\WhitePawnOnWhite}}}c3, is assigned a significantly larger piece value of -950 cp.
}
\label{fig3}
\end{figure}

Figure \ref{fig3} demonstrates that between two positions \cite{ref_vitiugov_ganguly,ref_nikitenko_mittal}, two pieces of the same color and type on the same square can have drastically different piece values. Our enhanced MLP+CNN piece value predictor assigns the two {\raisebox{-0.3ex}{\resizebox{9pt}{!}{\BlackBishopOnWhite}}}g7 significantly different values based on their relative activity and contribution to each position. Meanwhile, our MLP baselines are only able to assign a static value of -539 cp to each {\raisebox{-0.3ex}{\resizebox{9pt}{!}{\BlackBishopOnWhite}}}g7, reflecting the general principle that fianchettoed bishops are valuable whilst missing the key detail that their strength originates from their control over an open diagonal. This example showcases that utilizing piece location features alone for piece value predictions is insufficient, as they cannot capture the specific interactions between ally and enemy pieces which decide much of a position's evaluation.

\paragraph {Position representations.} The CNN position encoder uses 4, 6, or 8 convolutional layers with the following channel progressions:
\begin{itemize}
\item 4 layers: [32, 64, $d$, $d$]
\item 6 layers: [32, 64, 128, $d$, $d$, $d$]
\item 8 layers: [32, 64, 128, 256, $d$, $d$, $d$, $d$]
\end{itemize}
where $d$ is the representation dimension. Each convolutional layer applies a 2D convolution with a $3 \times 3$ kernel and padding of 1 to preserve intermediate spatial dimensions of $8 \times 8$, followed by ReLU activation and batch normalization. Early layers capture low-level spatial patterns such as piece adjacency while deeper layers encode more abstract positional features like the effect of different pawn structures on piece mobility. A final adaptive average pooling layer collapses the spatial dimensions, producing a $d$-dimensional representation vector.

The corresponding decoder mirrors this architecture, projecting the representation back to the initial $12 \times 8 \times 8$ board embedding through a linear expansion and a symmetric sequence of convolutional layers, with a sigmoid activation on the final layer to constrain outputs to $[0, 1]$. Preliminary tests of piece value predictors using position representation sizes of $d \in \{128, 256, 512\}$ showed that $d=512$ performed best, so all reported results use $d=512$.

\paragraph {MLP+CNN Predictors.} Nine MLP+CNN piece value predictor models were trained and evaluated, combining the three CNN position encoder depths with three MLP piece value predictor configurations of 3, 4, or 5 hidden layers:
\begin{itemize}
\item 3 layers: [256, 128, 64]
\item 4 layers: [512, 256, 128, 64]
\item 5 layers: [1024, 512, 256, 128, 64]
\end{itemize}
The MLP piece value predictor takes a 526-dimensional input formed by concatenating the 512-dim CNN-encoded position representation with the 14-dimensional piece feature vector (10-dim one-hot piece type encoding + 4-dim normalized location feature vector). Each hidden layer applies a linear transformation followed by batch normalization, ReLU activation, and graduated dropout, where wider early layers receive higher dropout rates (e.g. $p=0.4$ for the widest layer in the 5-layer variant) that decrease for narrower later layers (down to $p=0.1$)\footnote{For specifics, please refer to the documented source code found at \url{https://github.com/ethanjtang/PAWN}}. A final linear layer maps the last hidden layer's output to a single scalar, representing the predicted value of the input piece in the given position. As with the baseline MLPs, all MLP+CNN piece value predictors are trained using Huber loss ($\delta=1.0$).

\section{Datasets}
\label{sec:datasets}

\subsection{Data Collection}
\label{sec:datasets_collection}

Training and evaluating the models described in Section \ref{sec:sysarch} requires large datasets of chess positions paired with ground-truth piece values computed using Equation \ref{eq1}. We constructed two such datasets\footnote{We open-source both datasets at \url{https://huggingface.co/datasets/ethanjtang/PAWN-piece-value-datasets}} from Grandmaster-level games using the 2025 edition of the ChessBase Mega Database \cite{ref_chessbase_mega}, each offering a different distribution of positions:

\begin{enumerate}
\item Dataset MC: 6,925 games from former World Chess Champion GM Magnus Carlsen were used to gather 11,673,269 piece value entries from 549,410 unique positions.
\item Dataset TF: 7,656 games from all GM-level Classical games played in 2023 were used to gather 12,263,049 piece value entries from 533,540 unique positions.
\end{enumerate}

\begin{figure}[ht]
\centering
\includegraphics[width=0.50\textwidth]{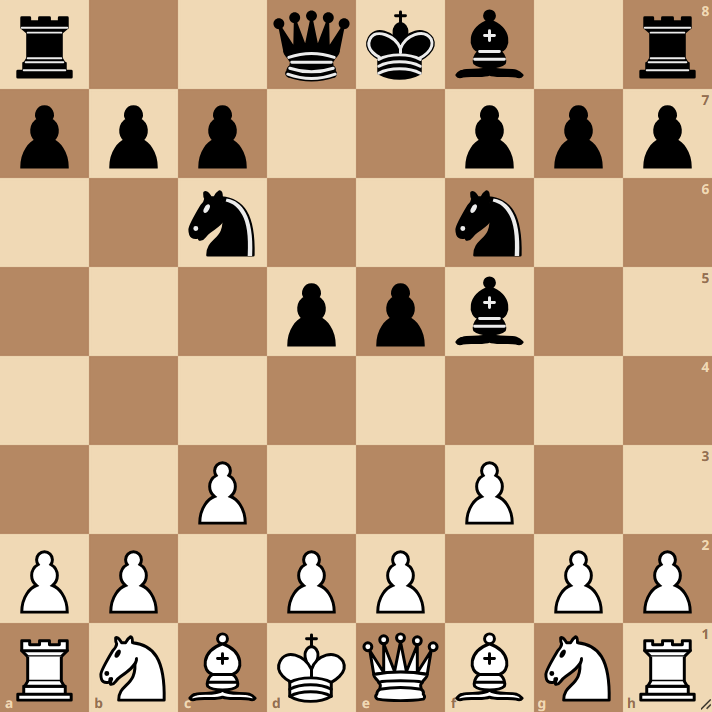}
\caption{A position in Dataset MC from one of GM Magnus Carlsen's online blitz games \cite{ref_carlsen_schneider}, where White (Carlsen) has chosen the non-standard plan of swapping his {\raisebox{-0.3ex}{\resizebox{9pt}{!}{\WhiteQueenOnWhite}}} and {\raisebox{-0.3ex}{\resizebox{9pt}{!}{\WhiteKingOnWhite}}} in the opening via {\raisebox{-0.3ex}{\resizebox{9pt}{!}{\WhiteQueenOnWhite}}}a4-h4-e1 and {\raisebox{-0.3ex}{\resizebox{9pt}{!}{\WhiteKingOnWhite}}}d1.
} 
\label{fig4}
\end{figure}

We train and evaluate models on both datasets independently.\footnote{We also open-source our best MLP and MLP+CNN models for each individual dataset at \\\url{https://huggingface.co/ethanjtang/PAWN-piece-value-predictors}} Dataset TF consists exclusively of Classical time format games played by GM-level players (both players above 2500 FIDE Classical Elo), where players have 2+ hours to make their moves and positions tend to follow established opening theory. Dataset MC includes a mix of time formats (ranging from 1-minute bullet games to multi-hour classical games) and both online and in-person games played by GM Magnus Carlsen. Comparing performance across both datasets allows us to examine how the distribution of positions in our training data affects piece value prediction accuracy, as faster time controls and online play tend to produce less conventional positions (see Fig. \ref{fig4}).

Ground-truth piece values were generated using Stockfish 17 at depth 20 with a timeout of 300 seconds per evaluation. Approximately 1\% of piece values across all processed positions could not be computed, either because removing the piece produced an illegal position (as described in Section \ref{sec:problem_formal_pval_def}) or due to non-deterministic behavior of Stockfish 17 when using multi-threaded, multi-core processing \cite{ref_czarnul,ref_sf_threading}.

\subsection{Preprocessing}
\label{sec:datasets_preprocess}

Before training, piece value entries were standardized using Z-score normalization based on the mean and standard deviation of the training set. We also experimented with capping piece values at five times their standard material values in centipawns ({\raisebox{-0.5ex}{\resizebox{10pt}{!}{\WhitePawnOnWhite}}}$\pm$500,
{\raisebox{-0.5ex}{\resizebox{10pt}{!}{\WhiteKnightOnWhite}}}$\pm$1500,
{\raisebox{-0.5ex}{\resizebox{10pt}{!}{\WhiteBishopOnWhite}}}$\pm$1500,
{\raisebox{-0.5ex}{\resizebox{10pt}{!}{\WhiteRookOnWhite}}}$\pm$2500,
{\raisebox{-0.5ex}{\resizebox{10pt}{!}{\WhiteQueenOnWhite}}}$\pm$5000 cp) to reduce the impact of extreme outliers. However, only approximately 1.6\% of piece values fell outside these caps, and capping had a minimal effect on prediction accuracy across all models ($\sim$3 cp improvement in the best case). We therefore omitted capping from our final pipeline, as the combination of Huber loss and Z-score normalization proved sufficient for handling extreme values. We discuss the effect of piece value capping in more detail in Appendix \ref{appendix_c_piece_capping}.

\subsection{Training and Validation Split}
\label{sec:datasets_trainval}

For each dataset, we use an 80:20 training:validation split at the game level, with all piece value entries from a single game assigned to a single set. Some position overlap remains ($\sim$6--8\% of validation positions also appear in the training data) due to common openings shared across different games. Initial iterations of our piece value predictor code utilized a row-level split, leading to significant overlap between the set of training and validation positions for our CNN position encoders. We discuss the progression of these experiments and the intermediate results that led us to adopt a game-level split in Appendix \ref{appendix_c_split_analysis}.

\section{Evaluations}
\label{sec:evals}

\subsection{Piece Value Predictor Results}
\label{sec:evals_pval_results}

Using the training and validation splits described in Section \ref{sec:datasets_trainval}, we train and evaluate both types of piece value predictors (MLP and MLP+CNN) independently on each dataset. CNN position encoders see only positions from the training split of their respective dataset. Performance is measured by mean absolute error $MAE$ in centipawns between predicted and ground-truth piece values:

\begin{equation}
\label{eq2}
\text{MAE} = \frac{1}{n} \sum_{i=1}^{n} |v_i - \hat{v}_i|
\end{equation}

\noindent where:

\begin{itemize}[leftmargin=2em]
\item[$n$] is the total number of piece values predicted
\item[$v_i$] is the ground-truth value for piece $i$ from Stockfish 17, as defined in Equation \ref{eq1}
\item[$\hat{v}_i$] is the corresponding predicted value for piece $i$
\end{itemize}

For readability, we report only the top-performing MLP+CNN configuration on each dataset (based on validation MAE), which was the 4-layer CNN encoder paired with a 5-layer MLP piece value predictor for both Dataset MC and TF. Detailed results for all 9 MLP+CNN configurations are included in Appendix \ref{appendix_b}.

\begin{table}[ht]
\caption{Model Performance Comparison for Dataset MC}\label{tab:model_performance_dataset_mc}
\centering
\setlength{\tabcolsep}{4pt}
\renewcommand{\arraystretch}{1.2}
\begin{tabular}{|l|c|c|}
\hline
\textbf{Model} & \textbf{Train MAE} & \textbf{Val MAE} \\
\hline
MLP \#1 (Gupta) & 84.52 cp & 83.74 cp \\
\hline
MLP \#2 & 84.00 cp & 83.19 cp \\
\hline
MLP \#3 & 84.19 cp & 83.40 cp \\
\hline
MLP+CNN & \textcolor{blue!80!black}{56.65 cp} & \textcolor{blue!80!black}{72.67 cp} \\
\hline
\end{tabular}
\end{table}

\begin{table}[ht]
\caption{Model Performance Comparison for Dataset TF}\label{tab:model_performance_dataset_tf}
\centering
\setlength{\tabcolsep}{4pt}
\renewcommand{\arraystretch}{1.2}
\begin{tabular}{|l|c|c|}
\hline
\textbf{Model} & \textbf{Train MAE} & \textbf{Val MAE} \\
\hline
MLP \#1 (Gupta) & 78.51 cp & 78.23 cp \\        
\hline
MLP \#2 & 78.26 cp & 77.99 cp \\        
\hline
MLP \#3 & 78.29 cp & 78.06 cp \\
\hline
MLP+CNN & \textcolor{blue!80!black}{52.87 cp} & \textcolor{blue!80!black}{65.45 cp} \\ 
\hline
\end{tabular}
\end{table}

Across both datasets, our three MLP baselines perform nearly identically, with validation MAE varying by less than 1 cp between them. Neither the additional hidden layer in MLP \#2 nor the non-linear piece location features in MLP \#3 (rank$^2$ and file$^2$) provide meaningful improvement over Gupta's original architecture, suggesting that increased model capacity and richer per-piece features alone are insufficient for capturing positional context.

Our enhanced MLP+CNN piece value predictor substantially outperforms all three MLP baselines on both datasets. We compare against MLP \#2 in our analysis, as it achieved the lowest validation MAE among the three baselines on both datasets. On Dataset MC, validation MAE drops from 83.19 cp to 72.67 cp, a reduction of 12.65\%. On Dataset TF, the improvement is larger: validation MAE falls from 77.99 cp to 65.45 cp, a reduction of 16.08\%. The stronger improvement on Dataset TF may reflect its more conventional distribution of positions, as Classical games between strong players exhibit more consistent structural patterns (pawn structures, opening trends, etc.), producing a more consistent training distribution for both the CNN position encoder and MLP piece value predictor.

Both datasets show a gap between training and validation MAE for the MLP+CNN predictor models (56.65 vs. 72.67 cp on Dataset MC, 52.87 vs. 65.45 cp on Dataset TF), indicating some degree of overfitting. However, this gap is notably smaller on Dataset TF, suggesting that the model generalizes more effectively from the structurally consistent positions in Classical games than from the diverse positions produced by Magnus Carlsen's varied time controls and playing contexts. Overall, our best-performing MLP+CNN configuration achieves a validation MAE of 65.45 cp on Dataset TF, predicting a piece's value within approximately 0.65 pawns.

\subsection{Analysis of Our Piece Value Definition}
\label{sec:evals_pval_def_application}

We also evaluate the application of our ablation-based piece value definition to derive a general piece valuation system. Table \ref{tab:piece_values} contains piece value statistics from Dataset TF, including piece type, mean value, median value, and total piece count. Because chess engines evaluate positions using positive values for White advantages and negative values for Black advantages \cite{ref_evaluation_wiki}, White pieces generally carry positive piece values while Black pieces generally carry negative values. By averaging the absolute median value of each piece type across both colors and normalizing by the pawn median, we extract the following general piece valuation system:

\begin{center}
{\raisebox{-0.3ex}{\resizebox{10pt}{!}{\WhitePawnOnWhite}}}=1, {\raisebox{-0.3ex}{\resizebox{10pt}{!}{\WhiteKnightOnWhite}}}=3.29, {\raisebox{-0.3ex}{\resizebox{10pt}{!}{\WhiteBishopOnWhite}}}=3.54, {\raisebox{-0.3ex}{\resizebox{10pt}{!}{\WhiteRookOnWhite}}}=3.77, {\raisebox{-0.3ex}{\resizebox{10pt}{!}{\WhiteQueenOnWhite}}}=5.14
\end{center}

Table~\ref{tab:valuation_comparison} compares our derived values with several established general piece valuation systems.

\begin{table}[ht]
\caption{Comparison of General Piece Valuation Systems}\label{tab:valuation_comparison}
\centering
\setlength{\tabcolsep}{4pt}
\renewcommand{\arraystretch}{1.25}
\begin{tabular}{|l|c|c|c|c|c|}
\hline
\textbf{System} & {\raisebox{-0.3ex}{\resizebox{9pt}{!}{\WhitePawnOnWhite}}} & {\raisebox{-0.3ex}{\resizebox{9pt}{!}{\WhiteKnightOnWhite}}} & {\raisebox{-0.3ex}{\resizebox{9pt}{!}{\WhiteBishopOnWhite}}} & {\raisebox{-0.3ex}{\resizebox{9pt}{!}{\WhiteRookOnWhite}}} & {\raisebox{-0.3ex}{\resizebox{9pt}{!}{\WhiteQueenOnWhite}}} \\
\hline
Traditional \cite{ref_piece_values_wiki} & 1 & 3 & 3 & 5 & 10 \\
\hline
Kaufman \cite{ref_kaufman_chesscom} & 1 & 3.25 & 3.5 & 5 & 9.75 \\
\hline
AlphaZero \cite{ref_alphazero_balance} & 1 & 3.05 & 3.33 & 5.63 & 9.5 \\
\hline
PAWN (ablation-based) & 1 & 3.29 & 3.54 & 3.77 & 5.14 \\
\hline
\end{tabular}
\end{table}

While our derived {\raisebox{-0.3ex}{\resizebox{10pt}{!}{\WhitePawnOnWhite}}}, {\raisebox{-0.3ex}{\resizebox{10pt}{!}{\WhiteKnightOnWhite}}}, and {\raisebox{-0.3ex}{\resizebox{10pt}{!}{\WhiteBishopOnWhite}}} values closely match Kaufman's 2022 system, both the {\raisebox{-0.3ex}{\resizebox{10pt}{!}{\WhiteRookOnWhite}}} and {\raisebox{-0.3ex}{\resizebox{10pt}{!}{\WhiteQueenOnWhite}}} appear to be substantially undervalued relative to all established systems. We hypothesize that this discrepancy arises from Stockfish's evaluation being calibrated to Win/Draw/Loss (WDL) probabilities rather than raw material count \cite{ref_sf_wdl} starting from version 15.1. As a result, removing material past a certain point has diminishing effects on the evaluation in most cases, leading our ablation-based definition to systematically undervalue rooks and queens. In simple terms, once a player has already lost a minor piece ({\raisebox{-0.3ex}{\resizebox{10pt}{!}{\WhiteKnightOnWhite}}}/{\raisebox{-0.3ex}{\resizebox{10pt}{!}{\WhiteBishopOnWhite}}}) in material, losing additional material beyond that point changes the expected game outcome less and less. 

\begin{table}[ht]
\caption{Piece Value Statistics from Dataset TF}\label{tab:piece_values}
\centering
\setlength{\tabcolsep}{4pt}
\renewcommand{\arraystretch}{1.25}
\begin{tabular}{|c|c|c|c|}
\hline
\textbf{Piece} & \textbf{Mean Val} & \textbf{Median Val} & \textbf{Count} \\
\hline
{\raisebox{-0.5ex}{\resizebox{10pt}{!}{\WhitePawnOnWhite}}} & $+156.39$ & $+147.00$ & 3,611,758 \\
\hline
{\raisebox{-0.5ex}{\resizebox{10pt}{!}{\WhiteKnightOnWhite}}} & $+500.62$ & $+507.00$ & 671,324 \\
\hline
{\raisebox{-0.5ex}{\resizebox{10pt}{!}{\WhiteBishopOnWhite}}} & $+538.49$ & $+545.00$ & 746,785 \\
\hline
{\raisebox{-0.5ex}{\resizebox{10pt}{!}{\WhiteRookOnWhite}}} & $+606.72$ & $+584.00$ & 713,663 \\
\hline
{\raisebox{-0.5ex}{\resizebox{10pt}{!}{\WhiteQueenOnWhite}}} & $+810.28$ & $+800.00$ & 418,339 \\
\hline
{\raisebox{-0.5ex}{\resizebox{10pt}{!}{\BlackPawnOnWhite}}} & $-167.85$ & $-160.00$ & 3,595,427 \\
\hline
{\raisebox{-0.5ex}{\resizebox{10pt}{!}{\BlackKnightOnWhite}}} & $-492.37$ & $-503.00$ & 678,542 \\
\hline
{\raisebox{-0.5ex}{\resizebox{10pt}{!}{\BlackBishopOnWhite}}} & $-528.59$ & $-541.00$ & 739,861 \\
\hline
{\raisebox{-0.5ex}{\resizebox{10pt}{!}{\BlackRookOnWhite}}} & $-590.16$ & $-573.00$ & 675,515 \\
\hline
{\raisebox{-0.5ex}{\resizebox{10pt}{!}{\BlackQueenOnWhite}}} & $-784.00$ & $-779.00$ & 411,835 \\
\hline
\end{tabular}
\end{table}

\clearpage 

\section{Conclusions}
\label{sec:conclusion}

\subsection{Implications}
\label{sec:conclusion_implications}

All three MLP baselines perform near-identically, varying by less than 1 cp in validation MAE despite differences in model capacity and input features. This suggests that the baseline (Color, Piece, Square) representation in \cite{ref_gupta} captures limited positional context and additional model complexity alone is unlikely to close the remaining gap. Incorporating an intermediate representation of the full board state derived from a CNN encoder yields a meaningful improvement over the baseline, reducing validation MAE by 16\% on Dataset TF to approximately 0.65 pawns of the Stockfish 17 ground truth. However, this level of error remains far from precise, indicating substantial room for improvement in both architecture and input representation.

While a comprehensive qualitative evaluation across all positions in our dataset is beyond the scope of this paper, the illustrative examples in Figures \ref{fig1}, \ref{fig3}, and \ref{fig5} demonstrate that our piece value predictor captures meaningful differences in piece quality, distinguishing between strong and weak pieces of the same type on the same square and assigning values consistent with well-understood positional principles.

More broadly, our findings suggest that predictive systems for individual component contributions can benefit from incorporating a vector representation of the entire problem state as context. In chess, a piece's value is determined not by its identity and location alone but also by its spatial relationships with every other piece on the board. The same motif appears in other domains: The value of an individual asset in a portfolio depends on its correlations with other holdings, and the meaning of a single word depends on the tokens that surround it. In each case, the ``whole'' shapes the value of each ``part'', and predictive systems that incorporate global context should outperform those that do not.

\subsection{Limitations}
\label{sec:conclusion_limits}

\paragraph{Ablation-based definition.} Our definition of piece value (Equation \ref{eq1}) measures the change in engine evaluation when a single piece is removed, providing a direct estimate of how much an individual piece contributes to a position. However, this treats each piece independently and ignores interaction effects, since removing one piece may change the effective value of others in the resulting position. 

\paragraph{Ground-truth dependence on Stockfish 17.} All piece value labels are derived from Stockfish 17 evaluations at depth 20. While Stockfish 17 is among the strongest publicly available engines, different versions/configurations of Stockfish or alternative engines may produce different piece values for the same position. Additionally, approximately 1\% of piece values could not be computed due to either illegal positions or non-deterministic behavior by Stockfish \cite{ref_czarnul,ref_sf_threading}, and we did not investigate whether this data loss introduces systematic bias.

\paragraph{Training-validation overlap.} Although we adopted a game-level split for our piece value data, approximately 6-8\% of validation positions also appear in the training data due to common openings shared across different games (Section \ref{sec:datasets_trainval}). This residual overlap may inflate MLP+CNN performance if the CNN position encoder partially memorizes frequently occurring board states.

\paragraph{Dataset-dependent accuracy.} The difference in validation MAE between Dataset MC and Dataset TF (72.67 cp vs. 65.45 cp) demonstrates sensitivity to the distribution of training positions. Deploying the model on positions from substantially different contexts may yield degraded performance without retraining, as piece value distributions likely differ across skill levels and time controls. For example, knights become more dangerous in faster time control games \cite[p.~31--32]{ref_silman}, where their tricky movement is harder to navigate under time pressure, shifting the relative value of pieces in ways that Grandmaster-level Classical games would not reflect.

\subsection{Future Work}
\label{sec:conclusion_future}

\paragraph{Enriched input piece features.} While our attempts at improving MLP piece value predictor performance via increased model capacity and richer per-piece features were unsuccessful, other features could add partial positional context without requiring a full board representation. Temporal features such as move number or game phase (via material count) could yield more accurate relative piece value predictions, as demonstrated by Kaufman's phase-dependent general piece valuation system \cite{ref_kaufman_chesscom,ref_kaufman_imbalances}. Other categorical features worth examining include opening type, relative strength of players, and time control.
\paragraph{Richer board representations.} The CNN autoencoders used in our MLP+CNN models encode only piece placement, omitting game-state information such as side to move, castling rights, and en passant availability. Of these, side to move affects the evaluation of every position and its omission may contribute to systematic prediction error. Future work should examine whether incorporating these state features into the board representation yields measurable improvements in piece value prediction.

\paragraph{Alternative architectures.} Graph neural networks (GNNs) offer a natural alternative to CNNs for encoding chess positions, as they can represent pieces as nodes and their spatial relationships as edges, potentially capturing long-range interactions more directly than convolutional filters on an $8 \times 8$ board. Transformer-based architectures are another promising direction, as recent work has shown that transformers can achieve comparable accuracy to Stockfish's NNUE-based evaluation for positions \cite{ref_chessformer}. Comparing CNN, GNN, and transformer-based position encoders for piece value prediction is a natural next step.

\paragraph{Downstream applications.} While our paper demonstrates that CNN-derived context improves piece value prediction, a gap remains between improved prediction accuracy and practical utility. Position evaluation functions could incorporate predicted piece quality as a supplementary feature. For human players, accurate piece value predictions could function as an interpretability tool, highlighting which pieces are performing above or below their general value and providing positional insights that raw engine evaluations do not offer.

\begin{credits}

\subsubsection{\ackname}
We thank Research Computing at Arizona State University \cite{ref_jennewein} for providing computing and storage resources for all experiments. Chess game data was sourced from the ChessBase Mega Database 2025 \cite{ref_chessbase_mega}.

\subsubsection{\discintname}
The authors have no competing interests to declare that are
relevant to the content of this article.

\end{credits}

\bibliographystyle{splncs04}
\bibliography{PAWN-references}

\clearpage
\appendix
\section{MLP+CNN Configuration Performance}
\label{appendix_a}

\noindent Table \ref{tab5:dataset_mc_aug_configs} showcases the performance of all MLP+CNN piece value predictor configurations trained on Dataset MC. Similarly, Table \ref{tab6:dataset_tf_aug_configs} showcases the performance of all MLP+CNN configurations trained on Dataset TF.

\begin{table}[ht]
\caption{Performance of all MLP+CNN Configurations on Dataset MC}
\label{tab5:dataset_mc_aug_configs}
\centering
\setlength{\tabcolsep}{4pt}
\renewcommand{\arraystretch}{1.2}
\begin{tabular}{|l|c|c|}
\hline
\textbf{Model} & \textbf{Train MAE} & \textbf{Val MAE} \\
\hline
4-layer CNN, 3-layer MLP &  67.70 cp & 75.36 cp \\
\hline
4-layer CNN, 4-layer MLP & 61.49 cp & 74.02 cp \\
\hline
4-layer CNN, 5-layer MLP & \textcolor{blue!80!black}{56.65 cp} & \textcolor{blue!80!black}{72.67 cp} \\
\hline
6-layer CNN, 3-layer MLP & 70.29 cp & 75.85 cp \\
\hline
6-layer CNN, 4-layer MLP & 66.84 cp & 75.00 cp \\
\hline
6-layer CNN, 5-layer MLP & 60.33 cp & 73.86 cp  \\
\hline
8-layer CNN, 3-layer MLP & 72.06 cp & 75.93 cp \\
\hline
8-layer CNN, 4-layer MLP & 69.04 cp & 75.42 cp \\
\hline
8-layer CNN, 5-layer MLP & 65.23 cp & 74.76 cp \\
\hline
\end{tabular}
\end{table}

\begin{table}[ht]
\caption{Performance of all MLP+CNN Configurations on Dataset TF}
\label{tab6:dataset_tf_aug_configs}
\centering
\setlength{\tabcolsep}{4pt}
\renewcommand{\arraystretch}{1.2}
\begin{tabular}{|l|c|c|}
\hline
\textbf{Model} & \textbf{Train MAE} & \textbf{Val MAE} \\
\hline
4-layer CNN, 3-layer MLP & 62.08 cp & 68.51 cp \\
\hline
4-layer CNN, 4-layer MLP & 57.18 cp & 66.82 cp \\
\hline
4-layer CNN, 5-layer MLP & \textcolor{blue!80!black}{52.87 cp} & \textcolor{blue!80!black}{65.45 cp} \\
\hline
6-layer CNN, 3-layer MLP & 62.35 cp & 68.69 cp \\
\hline
6-layer CNN, 4-layer MLP & 58.09 cp & 67.25 cp \\
\hline
6-layer CNN, 5-layer MLP & 53.73 cp & 66.01 cp \\
\hline
8-layer CNN, 3-layer MLP & 65.66 cp & 69.89 cp \\
\hline
8-layer CNN, 4-layer MLP & 61.03 cp & 68.17 cp \\
\hline
8-layer CNN, 5-layer MLP & 56.09 cp & 67.01 cp \\
\hline
\end{tabular}
\end{table}

\noindent For Dataset MC and TF, we find that using a configuration of a 4-layer CNN position encoder along with a 5-layer MLP piece value predictor performs the best (by validation error).

\clearpage
\section{Practical Applications of PAWN}
\label{appendix_b}

\begin{figure}[ht]
\centering
\includegraphics[width=0.65\textwidth]{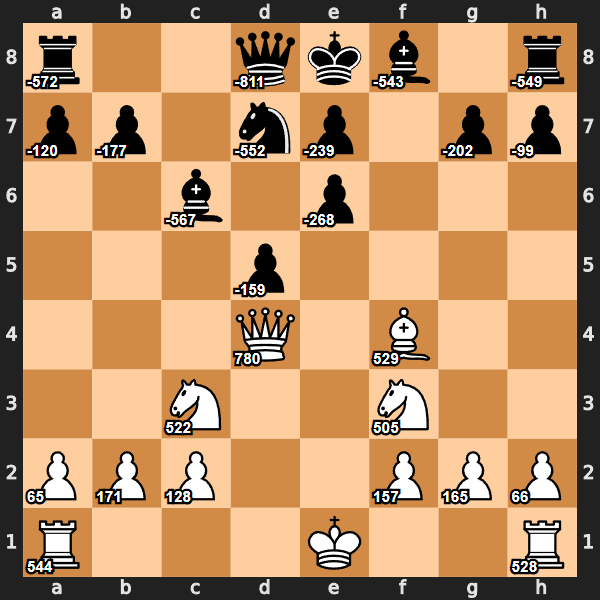}
\caption{Predicted piece values in the Jobava-Rapport system after 3..c5 4. e4! cxd4 ... 9.e6! fxe6 \cite{ref_jobava_london,ref_lichess_masters}.}
\label{fig5}
\end{figure}

As an example of a potential practical application for our piece value predictor, we also provide in-depth analysis of the following position (Fig. \ref{fig5}) with all piece values displayed in the bottom-left corner of each occupied square.

In this position, White has sacrificed a pawn on e6 in exchange for a lead in development. Play in this position revolves around the question of whether Black can finish their development before White coordinates an attack against Black's backwards {\raisebox{-0.3ex}{\resizebox{10pt}{!}{\BlackPawnOnWhite}}}e6/{\raisebox{-0.3ex}{\resizebox{10pt}{!}{\BlackPawnOnWhite}}}e7. The best move in the position is 10\ldots{\raisebox{-0.3ex}{\resizebox{10pt}{!}{\BlackQueenOnWhite}}}b6, offering a trade of queens that White declines with 11. {\raisebox{-0.3ex}{\resizebox{10pt}{!}{\WhiteQueenOnWhite}}}d2. Black then has the choice between 11\ldots{\raisebox{-0.3ex}{\resizebox{10pt}{!}{\BlackPawnOnWhite}}}d4 or 11\ldots{\raisebox{-0.3ex}{\resizebox{10pt}{!}{\BlackPawnOnWhite}}}e5, returning the extra pawn while reducing White's lead in development by trading off some of White's active pieces.
There are a few patterns present in this position that we note match up with conventional chess knowledge.

\begin{enumerate}
\item
White's {\raisebox{-0.3ex}{\resizebox{10pt}{!}{\WhitePawnOnWhite}}}a2/{\raisebox{-0.3ex}{\resizebox{10pt}{!}{\WhitePawnOnWhite}}}h2 are worth significantly less than any other White pawns. White's advantage is largely dynamic in this position due to their lead in development, so removing the {\raisebox{-0.3ex}{\resizebox{10pt}{!}{\WhitePawnOnWhite}}}a2/{\raisebox{-0.3ex}{\resizebox{10pt}{!}{\WhitePawnOnWhite}}}h2 does not heavily impact the evaluation of the position due to their removal activating either the {\raisebox{-0.3ex}{\resizebox{10pt}{!}{\WhiteRookOnWhite}}}a1/{\raisebox{-0.3ex}{\resizebox{10pt}{!}{\WhiteRookOnWhite}}}h1, allowing them to pressure the {\raisebox{-0.3ex}{\resizebox{10pt}{!}{\BlackPawnOnWhite}}}a7/{\raisebox{-0.3ex}{\resizebox{10pt}{!}{\BlackPawnOnWhite}}}h7 respectively.
\item
Black's pawns are worth more on average than White's. This is due to Black's advantage being largely static (up +1{\raisebox{-0.3ex}{\resizebox{10pt}{!}{\BlackPawnOnWhite}}}); losing any material would swing the position's evaluation heavily in White's favor due to White's advantage being primarily dynamic and therefore not as dependent (see Section \ref{sec:evals_pval_def_application}) on the static factor of material count.
\item
The {\raisebox{-0.3ex}{\resizebox{10pt}{!}{\BlackBishopOnWhite}}}c6 is the most valuable minor piece in this position, even outvaluing the {\raisebox{-0.3ex}{\resizebox{10pt}{!}{\BlackRookOnWhite}}}h8/{\raisebox{-0.3ex}{\resizebox{10pt}{!}{\WhiteRookOnWhite}}}a1/{\raisebox{-0.3ex}{\resizebox{10pt}{!}{\WhiteRookOnWhite}}}h1 despite it being understood that bishops are worth less than rooks in general valuation systems. The {\raisebox{-0.3ex}{\resizebox{10pt}{!}{\BlackBishopOnWhite}}}c6 prevents White's thematic idea of {\raisebox{-0.3ex}{\resizebox{10pt}{!}{\WhiteKnightOnWhite}}}b5-{\raisebox{-0.3ex}{\resizebox{10pt}{!}{\WhiteKnightOnWhite}}}c7 while all rooks in this position are inactive, leading to the {\raisebox{-0.3ex}{\resizebox{10pt}{!}{\BlackBishopOnWhite}}}c6 being valued more highly in this position due to the important role it serves.
\end{enumerate}

\clearpage
\section{PAWN Architecture Iterations}
\label{appendix_c}

This appendix documents the iterative experiments that shaped our final piece value predictor architecture. Section~\ref{appendix_c_arch_scaling} describes the cumulative architectural changes between our initial and final models and demonstrates their impact on both accuracy and generalization. Sections~\ref{appendix_c_embedding_dims}--\ref{appendix_c_piece_capping} then present three targeted experiments that informed specific design decisions carried forward into our final pipeline: position representation dimensionality, piece value data splitting, and piece value outlier handling. Although some of these experiments were conducted using our old architecture and smaller datasets, the findings transferred directly to our final models.

\subsection{Architectural and Dataset Scaling Improvements}
\label{appendix_c_arch_scaling}

Our final architecture differs from our initial system in five key ways:
 
\begin{enumerate}
    \item \textbf{Loss function:} Mean Squared Error (MSE) Loss $\rightarrow$ Huber Loss ($\delta = 1.0$). MSE disproportionately penalizes inaccurate predictions for extreme piece values, causing the model to optimize for outliers at the expense of typical piece values. Huber loss applies a linear penalty beyond the threshold $\delta$, reducing sensitivity to extreme values while preserving gradient signal for normal piece value predictions.
    \item \textbf{Target normalization:} Min-max normalization ($[0,1]$) $\rightarrow$ Z-score standardization. Min-max normalization used in our old architecture compressed the loss signal for outliers into a narrow range. Meanwhile, Z-score standardization allows unbounded output, enabling the model to better predict both normal and extreme piece values.
    \item \textbf{Weight decay:} Enabled weight decay in the AdamW optimizer ($\lambda$ > 0). Adding weight decay penalizes large parameter values, reducing overfitting in the MLP+CNN models.
    \item \textbf{Batch normalization:} Added to all hidden layers in both MLP and MLP+CNN predictors, stabilizing training dynamics.
    \item \textbf{Dropout:} Uniform dropout $\rightarrow$ graduated dropout for the MLP piece value predictor in augmented MLP+CNN models. Wider early layers receive higher dropout rates (e.g., $p = 0.4$) that decrease for narrower later layers (down to $p = 0.1$), applying stronger regularization where the model has the most capacity to overfit.
\end{enumerate}

We first isolate the impact of these architectural changes using Dataset TF, which remained unchanged between our initial and final experiments. Table~\ref{tab:dataset_tf_arch_comparison} compares the old and new architectures trained on the same data. For MLP+CNN models, the average training/validation gap decreases from 23.84~cp to 8.75~cp, a reduction of 63\%. Meanwhile, MLP baseline MAE drops from $\sim$132~cp to $\sim$78~cp, likely driven by the switch to Z-score standardization and Huber loss since MLP baselines do not use the CNN-specific BatchNorm or graduated dropout. While the architectural changes did not significantly affect the train/val MAE gap for MLP baselines (which remains near zero in both architectures), the substantial improvement in absolute MAE for all model types suggests that our optimizations broadly improve training quality.

\begin{table}[ht]
\caption{Performance of MLP and MLP+CNN Models on Dataset TF (Old vs. New Architecture)}
\label{tab:dataset_tf_arch_comparison}
\centering
\setlength{\tabcolsep}{4pt}
\renewcommand{\arraystretch}{1.2}
\begin{tabular}{|l|c|c|c|c|c|c|}
\hline
 & \multicolumn{3}{c|}{\textbf{Old Arch.}} & \multicolumn{3}{c|}{\textbf{New Arch.}} \\
\cline{2-7}
\textbf{Model} & \textbf{Train MAE} & \textbf{Val MAE} & \textbf{Gap} & \textbf{Train MAE} & \textbf{Val MAE} & \textbf{Gap} \\
\hline
\multicolumn{7}{|l|}{\textit{MLP Baselines}} \\
\hline
MLP \#1 (12-dim, 2 layers) & 132.99 cp & 132.82 cp & $-$0.17 & 78.51 cp & 78.23 cp & $-$0.28 \\
\hline
MLP \#2 (12-dim, 3 layers) & 132.66 cp & 132.52 cp & $-$0.14 & 78.26 cp & 77.99 cp & $-$0.27 \\
\hline
MLP \#3 (14-dim, 3 layers) & 132.75 cp & 132.64 cp & $-$0.11 & 78.29 cp & 78.06 cp & $-$0.23 \\
\hline
\multicolumn{7}{|l|}{\textit{MLP+CNN Models}} \\
\hline
4-layer CNN, 3-layer MLP & 106.01 cp & 124.93 cp & $+$18.92 & 62.08 cp & 68.51 cp & $+$6.43 \\
\hline
4-layer CNN, 4-layer MLP & 98.05 cp & 125.08 cp & $+$27.03 & 57.18 cp & 66.82 cp & $+$9.64 \\
\hline
4-layer CNN, 5-layer MLP & \textcolor{blue!80!black}{87.38 cp} & \textcolor{blue!80!black}{123.70 cp} & $+$36.32 & \textcolor{blue!80!black}{52.87 cp} & \textcolor{blue!80!black}{65.45 cp} & $+$12.58 \\
\hline
6-layer CNN, 3-layer MLP & 103.41 cp & 124.57 cp & $+$21.16 & 62.35 cp & 68.69 cp & $+$6.34 \\
\hline
6-layer CNN, 4-layer MLP & 112.28 cp & 124.40 cp & $+$12.12 & 58.09 cp & 67.25 cp & $+$9.16 \\
\hline
6-layer CNN, 5-layer MLP & 95.37 cp & 125.22 cp & $+$29.85 & 53.73 cp & 66.01 cp & $+$12.28 \\
\hline
8-layer CNN, 3-layer MLP & 110.04 cp & 125.48 cp & $+$15.44 & 65.66 cp & 69.89 cp & $+$4.23 \\
\hline
8-layer CNN, 4-layer MLP & 99.81 cp & 125.70 cp & $+$25.89 & 61.03 cp & 68.17 cp & $+$7.14 \\
\hline
8-layer CNN, 5-layer MLP & 98.03 cp & 125.84 cp & $+$27.81 & 56.09 cp & 67.01 cp & $+$10.92 \\
\hline
\end{tabular}
\end{table}

Table~\ref{tab:dataset_mc_size_comparison} presents a second comparison across both architecture and dataset scale simultaneously due to our Dataset MC-large (used in the main paper) being gathered in parallel with our architectural improvements. Dataset MC-small (not used in the main paper) was constructed using a smaller sample of 2,108 Classical games played by GM Magnus Carlsen, yielding 1,436,034 piece values from 160,183 unique positions, compared to the 11,673,269 piece values in MC-large. Despite the 8$\times$ increase in dataset size, the average gap between training and validation MAE for MLP+CNN models remains comparable, decreasing slightly from 11.53~cp to 9.25~cp. This suggests that our new piece value prediction architecture scales effectively to larger datasets without additional overfitting.

\begin{table}[ht]
\caption{Performance of MLP and MLP+CNN Models Under Combined Architecture and Dataset Changes (Old Architecture with MC-small vs. New Architecture with MC-large)}
\label{tab:dataset_mc_size_comparison}
\centering
\setlength{\tabcolsep}{4pt}
\renewcommand{\arraystretch}{1.2}
\begin{tabular}{|l|c|c|c|c|c|c|}
\hline
 & \multicolumn{3}{c|}{\textbf{Old Arch. + MC-small}} & \multicolumn{3}{c|}{\textbf{New Arch. + MC-large}} \\
\cline{2-7}
\textbf{Model} & \textbf{Train MAE} & \textbf{Val MAE} & \textbf{Gap} & \textbf{Train MAE} & \textbf{Val MAE} & \textbf{Gap} \\
\hline
\multicolumn{7}{|l|}{\textit{MLP Baselines}} \\
\hline
MLP \#1 (12-dim, 2 layers) & 82.42 cp & 82.16 cp & $-$0.26 & 84.00 cp & 83.19 cp & $-$0.81 \\
\hline
MLP \#2 (12-dim, 3 layers) & 85.18 cp & 84.86 cp & $-$0.32 & 84.19 cp & 83.40 cp & $-$0.79 \\
\hline
MLP \#3 (14-dim, 3 layers) & 87.03 cp & 86.68 cp & $-$0.35 & 84.52 cp & 83.74 cp & $-$0.78 \\
\hline
\multicolumn{7}{|l|}{\textit{MLP+CNN Models}} \\
\hline
4-layer CNN, 3-layer MLP & 56.93 cp & 64.47 cp & $+$7.54 & 67.70 cp & 75.36 cp & +7.66 \\
\hline
4-layer CNN, 4-layer MLP & 46.55 cp & 60.12 cp & $+$13.57 & 61.49 cp & 74.02 cp & +12.53 \\
\hline
4-layer CNN, 5-layer MLP & 44.25 cp & \textcolor{blue!80!black}{59.73 cp} & $+$15.48 & \textcolor{blue!80!black}{56.65 cp} & \textcolor{blue!80!black}{72.67 cp} & +16.02 \\
\hline
6-layer CNN, 3-layer MLP & 55.69 cp & 63.81 cp & $+$8.12 & 70.29 cp & 75.85 cp & +5.56 \\
\hline
6-layer CNN, 4-layer MLP & 51.13 cp & 61.86 cp & $+$10.73 & 66.84 cp & 75.00 cp & +8.16 \\
\hline
6-layer CNN, 5-layer MLP & \textcolor{blue!80!black}{43.43 cp} & 60.76 cp & $+$17.33 & 60.33 cp & 73.86 cp & +13.53 \\
\hline
8-layer CNN, 3-layer MLP & 54.51 cp & 63.42 cp & $+$8.91 & 72.06 cp & 75.93 cp & +3.87 \\
\hline
8-layer CNN, 4-layer MLP & 55.06 cp & 65.78 cp & $+$10.72 & 69.04 cp & 75.42 cp & +6.38 \\
\hline
8-layer CNN, 5-layer MLP & 54.86 cp & 66.24 cp & $+$11.38 & 65.23 cp & 74.76 cp & +9.53 \\
\hline
\end{tabular}
\end{table}

\clearpage

\subsection{Why use d=512?}
\label{appendix_c_embedding_dims}

To determine the optimal size for our $d$-dimensional CNN-encoded position representations, we compared three configurations ($d \in \{128, 256, 512\}$) using Dataset MC-small with our old architecture. Although these experiments predate the architectural improvements described in Section~\ref{appendix_c_arch_scaling}, the relative ranking of representation dimensions informed our choice of $d = 512$ for all final MLP+CNN configurations.

\begin{table}[ht]
\caption{Performance of CNN Configurations with Varying Representation Dimensions on Dataset MC-small}
\label{tab:dataset_mc_small_embedding_dims}
\centering
\setlength{\tabcolsep}{4pt}
\renewcommand{\arraystretch}{1.2}
\begin{tabular}{|l|c|c|c|}
\hline
\textbf{Model} & \textbf{Train MAE} & \textbf{Val MAE} & \textbf{Gap} \\
\hline
\multicolumn{4}{|l|}{\textit{$d = 128$}} \\
\hline
4-layer CNN, 128d & 52.87 cp & 63.44 cp & $+$10.57 \\
\hline
6-layer CNN, 128d & 59.86 cp & 67.00 cp & $+$7.14 \\
\hline
8-layer CNN, 128d & 56.30 cp & 64.91 cp & $+$8.61 \\
\hline
\multicolumn{4}{|l|}{\textit{$d = 256$}} \\
\hline
4-layer CNN, 256d & 49.98 cp & 61.12 cp & $+$11.14 \\
\hline
6-layer CNN, 256d & 51.65 cp & 62.30 cp & $+$10.65 \\
\hline
8-layer CNN, 256d & 52.78 cp & 63.66 cp & $+$10.88 \\
\hline
\multicolumn{4}{|l|}{\textit{$d = 512$}} \\
\hline
4-layer CNN, 512d & \textcolor{blue!80!black}{47.47 cp} & \textcolor{blue!80!black}{59.97 cp} & $+$12.50 \\
\hline
6-layer CNN, 512d & 51.86 cp & 62.04 cp & $+$10.18 \\
\hline
8-layer CNN, 512d & 69.65 cp & 74.29 cp & $+$4.64 \\
\hline
\end{tabular}
\end{table}

All configurations in Table~\ref{tab:dataset_mc_small_embedding_dims} use the same 3-layer MLP piece value predictor with hidden layers of  $[256, 128, 64]$, ReLU activations, and dropout ($p=0.1$). The MLP piece value predictor input consists of the concatenation of the $d$-dimensional CNN representation with the 14-dimensional piece feature vector.

The 4-layer CNN encoder with $d = 512$ achieves the lowest validation MAE (59.97~cp) across all configurations. We note additionally that deeper CNN encoders did not benefit from larger representation dimensions, with the 8-layer $d = 512$ configuration performing worst overall. Based on these results, we adopted $d = 512$ with our best-performing 4-layer CNN encoder for all reported experiments in the main paper.

\subsection{Row vs. Game-level Split for Piece Value Data}
\label{appendix_c_split_analysis}

We tested whether splitting piece value data by game rather than by individual row would reduce overfitting in our MLP+CNN models. Under a row-level split, positions from the same game can appear in both the training and validation sets, allowing the CNN position encoder to encounter similar or identical board states across both sets. A game-level split assigns all positions from each game exclusively to one set, reducing this source of data leakage (as noted previously in Section \ref{sec:datasets_trainval}, train/val position sets are never disjoint due to shared openings between games).

\clearpage
\begin{table}[ht]
\caption{Performance of Old Architecture on Dataset MC-small (Row-level vs Game-level Split)}
\label{tab:dataset_mc_small_split_comparison}
\centering
\setlength{\tabcolsep}{4pt}
\renewcommand{\arraystretch}{1.2}
\begin{tabular}{|l|c|c|c|c|c|c|}
\hline
 & \multicolumn{3}{c|}{\textbf{Row-level Split}} & \multicolumn{3}{c|}{\textbf{Game-level Split}} \\
\cline{2-7}
\textbf{Model} & \textbf{Train MAE} & \textbf{Val MAE} & \textbf{Gap} & \textbf{Train MAE} & \textbf{Val MAE} & \textbf{Gap} \\
\hline
\multicolumn{7}{|l|}{\textit{MLP Baselines}} \\
\hline
MLP \#1 (12-dim, 2 layers) & 132.99 cp & 132.82 cp & $-$0.17 & 132.42 cp & 131.77 cp & $-$0.65 \\
\hline
MLP \#2 (12-dim, 3 layers) & 132.66 cp & 132.52 cp & $-$0.14 & 132.55 cp & 131.94 cp & $-$0.61 \\
\hline
MLP \#3 (14-dim, 3 layers) & 132.75 cp & 132.64 cp & $-$0.11 & 133.47 cp & 132.82 cp & $-$0.65 \\
\hline
\multicolumn{7}{|l|}{\textit{MLP+CNN Models}} \\
\hline
4-layer CNN, 3-layer MLP & 106.01 cp & 124.93 cp & $+$18.92 & 108.47 cp & 123.06 cp & $+$14.59 \\
\hline
4-layer CNN, 4-layer MLP & 98.05 cp & 125.08 cp & $+$27.03 & 100.98 cp & 122.12 cp & $+$21.14 \\
\hline
4-layer CNN, 5-layer MLP & \textcolor{blue!80!black}{87.38 cp} & \textcolor{blue!80!black}{123.70 cp} & $+$36.32 & \textcolor{blue!80!black}{92.62 cp} & \textcolor{blue!80!black}{121.94 cp} & $+$29.32 \\
\hline
6-layer CNN, 3-layer MLP & 103.41 cp & 124.57 cp & $+$21.16 & 109.18 cp & 123.85 cp & $+$14.67 \\
\hline
6-layer CNN, 4-layer MLP & 112.28 cp & 124.40 cp & $+$12.12 & 105.52 cp & 123.02 cp & $+$17.50 \\
\hline
6-layer CNN, 5-layer MLP & 95.37 cp & 125.22 cp & $+$29.85 & 94.22 cp & 122.94 cp & $+$28.72 \\
\hline
8-layer CNN, 3-layer MLP & 110.04 cp & 125.48 cp & $+$15.44 & 118.77 cp & 124.79 cp & $+$6.02 \\
\hline
8-layer CNN, 4-layer MLP & 99.81 cp & 125.70 cp & $+$25.89 & 109.34 cp & 124.76 cp & $+$15.42 \\
\hline
8-layer CNN, 5-layer MLP & 98.03 cp & 125.84 cp & $+$27.81 & 104.33 cp & 125.07 cp & $+$20.74 \\
\hline
\end{tabular}
\end{table}

Table~\ref{tab:dataset_mc_small_split_comparison} shows that using a game-level split on our piece value data has a limited direct impact on validation MAE. For MLP+CNN models, validation MAE decreases by an average of approximately 2~cp compared to the row-level split. However, training MAE increases for 7/9 model configurations under the game split, indicating that the game-level split successfully reduces memorization of position-specific patterns without degrading the model's ability to generalize. We adopted the game-level split for all final experiments, as described in Section \ref{sec:datasets_trainval} of the main paper.

\subsection{Why not use piece value capping for outliers?}
\label{appendix_c_piece_capping}

We investigated whether capping extreme piece values at five times their standard material values in centipawns ({\raisebox{-0.5ex}{\resizebox{10pt}{!}{\WhitePawnOnWhite}}}$\pm$500,
{\raisebox{-0.5ex}{\resizebox{10pt}{!}{\WhiteKnightOnWhite}}}$\pm$1500,
{\raisebox{-0.5ex}{\resizebox{10pt}{!}{\WhiteBishopOnWhite}}}$\pm$1500,
{\raisebox{-0.5ex}{\resizebox{10pt}{!}{\WhiteRookOnWhite}}}$\pm$2500,
{\raisebox{-0.5ex}{\resizebox{10pt}{!}{\WhiteQueenOnWhite}}}$\pm$5000 cp) would improve prediction accuracy. Table~\ref{tab:capping_stats} showcases that only 1.6\% of piece values in Dataset MC-large exceed these thresholds, with nearly all affected entries belonging to pawns.

\begin{table}[ht]
\caption{Piece Value Capping Statistics on Dataset MC-large}\label{tab:capping_stats}
\centering
\setlength{\tabcolsep}{4pt}
\renewcommand{\arraystretch}{1.25}
\begin{tabular}{|c|c|c|c|}
\hline
\textbf{Piece} & \textbf{Cap Threshold} & \textbf{Values Capped} & \textbf{\% of Total Rows} \\
\hline
{\raisebox{-0.5ex}{\resizebox{10pt}{!}{\WhitePawnOnWhite}}} & $\pm$500 cp & 186,423 & 1.5970\% \\
\hline
{\raisebox{-0.5ex}{\resizebox{10pt}{!}{\WhiteKnightOnWhite}}} & $\pm$1,500 cp & 313 & 0.0027\% \\
\hline
{\raisebox{-0.5ex}{\resizebox{10pt}{!}{\WhiteBishopOnWhite}}} & $\pm$1,500 cp & 607 & 0.0052\% \\
\hline
{\raisebox{-0.5ex}{\resizebox{10pt}{!}{\WhiteRookOnWhite}}} & $\pm$2,500 cp & 10 & 0.0001\% \\
\hline
{\raisebox{-0.5ex}{\resizebox{10pt}{!}{\WhiteQueenOnWhite}}} & $\pm$5,000 cp & 1 & $<$0.0001\% \\
\hline
\textbf{Total} & -- & \textbf{187,354} & \textbf{1.60\%} \\
\hline
\end{tabular}
\end{table}

Table~\ref{tab:dataset_mc_large_capping} compares model performance with and without capping on Dataset MC-large using our final architecture. Capping produces a modest improvement of approximately 2--3~cp in validation MAE for the best MLP+CNN model (70.17~cp vs.\ 72.67~cp), with the improvement remaining consistent across all configurations. Despite this, we omitted capping from our final pipeline for two reasons: first, the improvement is small relative to the overall error; second, the combination of Huber loss and Z-score standardization already mitigates the influence of extreme values without discarding information.

\begin{table}[ht]
\caption{Performance of New Architecture on Dataset MC-large (Capping vs No Capping)}
\label{tab:dataset_mc_large_capping}
\centering
\setlength{\tabcolsep}{4pt}
\renewcommand{\arraystretch}{1.2}
\begin{tabular}{|l|c|c|c|c|c|c|}
\hline
 & \multicolumn{3}{c|}{\textbf{With Capping}} & \multicolumn{3}{c|}{\textbf{No Capping}} \\
\cline{2-7}
\textbf{Model} & \textbf{Train MAE} & \textbf{Val MAE} & \textbf{Gap} & \textbf{Train MAE} & \textbf{Val MAE} & \textbf{Gap} \\
\hline
\multicolumn{7}{|l|}{\textit{MLP Baselines}} \\
\hline
MLP \#1 (12-dim, 2 layers) & 81.57 cp & 80.87 cp & $-$0.70 & 84.00 cp & 83.19 cp & $-$0.81 \\
\hline
MLP \#2 (12-dim, 3 layers) & 81.08 cp & 80.37 cp & $-$0.71 & 84.19 cp & 83.40 cp & $-$0.79 \\
\hline
MLP \#3 (14-dim, 3 layers) & 81.23 cp & 80.54 cp & $-$0.69 & 84.52 cp & 83.74 cp & $-$0.78 \\
\hline
\multicolumn{7}{|l|}{\textit{MLP+CNN Models}} \\
\hline
4-layer CNN, 3-layer MLP & 64.63 cp & 72.53 cp & $+$7.90 & 67.70 cp & 75.36 cp & $+$7.66 \\
\hline
4-layer CNN, 4-layer MLP & 60.12 cp & 71.17 cp & $+$11.05 & 61.49 cp & 74.02 cp & $+$12.53 \\
\hline
4-layer CNN, 5-layer MLP & \textcolor{blue!80!black}{54.93 cp} & \textcolor{blue!80!black}{70.17 cp} & $+$15.24 & \textcolor{blue!80!black}{56.65 cp} & \textcolor{blue!80!black}{72.67 cp} & $+$16.02 \\
\hline
6-layer CNN, 3-layer MLP & 67.55 cp & 73.32 cp & $+$5.77 & 70.29 cp & 75.85 cp & $+$5.56 \\
\hline
6-layer CNN, 4-layer MLP & 63.40 cp & 72.13 cp & $+$8.73 & 66.84 cp & 75.00 cp & $+$8.16 \\
\hline
6-layer CNN, 5-layer MLP & 56.98 cp & 71.01 cp & $+$14.03 & 60.33 cp & 73.86 cp & $+$13.53 \\
\hline
8-layer CNN, 3-layer MLP & 68.32 cp & 73.55 cp & $+$5.23 & 72.06 cp & 75.93 cp & $+$3.87 \\
\hline
8-layer CNN, 4-layer MLP & 64.47 cp & 72.69 cp & $+$8.22 & 69.04 cp & 75.42 cp & $+$6.38 \\
\hline
8-layer CNN, 5-layer MLP & 59.34 cp & 71.62 cp & $+$12.28 & 65.23 cp & 74.76 cp & $+$9.53 \\
\hline
\end{tabular}
\end{table}

\subsection{Conclusions}

The experiments outlined in this appendix demonstrate that the primary source of improved accuracy and generalization in our final piece value predictors is the set of architectural changes described in Section~\ref{appendix_c_arch_scaling}, rather than our data splitting or outlier handling strategies outlined in Sections \ref{appendix_c_split_analysis} and \ref{appendix_c_piece_capping}. Among these changes, the switch from MSE to Huber loss combined with Z-score standardization appears to be the most impactful pair. Our MLP baselines, which do not benefit from graduated dropout or CNN-specific BatchNorm, still improve dramatically on Dataset TF (from $\sim$132~cp to $\sim$78~cp under the new architecture), indicating that the loss function and normalization changes alone account for a substantial portion of the gains. The remaining changes, namely AdamW with weight decay, BatchNorm, and graduated dropout, primarily benefit the MLP+CNN models by reducing the training/validation gap of each configuration from an average of 23.84~cp to 8.75~cp on Dataset TF.

We note that a gap between training and validation MAE persists in our final MLP+CNN results. Due to time and compute constraints, we were unable to isolate the contribution of each optimization individually or implement further improvements to our pipeline. Future work should balance improvements in piece value prediction accuracy with generalization to out-of-distribution applications.

\end{document}